\documentclass[11pt,letterpaper]{article}
\usepackage{ijcnlp2017}
\usepackage{times}
\usepackage{url}
\usepackage{amsmath,amssymb,graphicx,caption,bm}
\usepackage{algorithm,algorithmic,color,multirow,float,subcaption}
\usepackage[toc,page]{appendix}

\DeclareMathOperator{\bbR}{\mathbb{R}}
\DeclareMathOperator{\bbP}{\mathbb{P}}
\DeclareMathOperator{\bbN}{\mathbb{N}}
\DeclareMathOperator{\mW}{\mathbf{W}}

\DeclareMathOperator{\mg}{\bm{\Phi}}
\DeclareMathOperator{\mb}{\bm{\phi}}
\DeclareMathOperator{\vphi}{\mb}
\DeclareMathOperator{\err}{err}
\DeclareMathOperator{\vPhi}{\mg}
\DeclareMathOperator{\vy}{\mathbf{y}}
\DeclareMathOperator{\vtheta}{\bm{\theta}}
\DeclareMathOperator{\Cp}{C_{+}}
\DeclareMathOperator{\Cm}{C}
\DeclareMathOperator{\np}{N_{+}}
\DeclareMathOperator{\nm}{N}
\DeclareMathOperator{\vw}{\mathbf{w}}

\newcommand{\defeq}{\mbox{$\;\stackrel{\mbox{\tiny\rm def}}{=}\;$}}

\ijcnlpfinalcopy

\title{Sentiment Analysis by Joint Learning of Word Embeddings and Classifier}
\author{Prathusha Kameswar Sarma\and Bill Sethares \\
University of Wisconsin - Madison}
\date{}
\begin{document}
\maketitle
\begin{abstract}
Word embeddings are representations of individual words of a text document in a vector space and they are often useful for performing natural language processing tasks. Current state of the art algorithms for learning word embeddings learn vector representations from large corpora of text documents in an unsupervised fashion. This paper introduces SWESA (Supervised Word Embeddings for Sentiment Analysis), an algorithm for sentiment analysis via word embeddings. SWESA leverages document label information to learn vector representations of words from a modest corpus of text documents by solving an optimization problem that minimizes a cost function with respect to both word embeddings as well as classification accuracy.
Analysis reveals that SWESA provides an efficient way of estimating the dimension of the word embeddings that are to be learned. Experiments on several real world data sets 
show that SWESA has superior performance when compared to previously suggested approaches to word embeddings and sentiment analysis tasks.
\end{abstract}

\section{Introduction}
Representing words in a vector space allows quantification of relationships among words using distance or angle measures. Such vector representations for words are useful in performing several Natural Language Processing (NLP) tasks. The general idea when learning word embeddings is to estimate the underlying probability distribution function of a word from a given corpus of text documents. Most probabilistic models for learning semantic word embeddings, of which neural network based models such as the current state of the art word2vec algorithm and its derivatives~\cite{mikolov2013distributed,le2014distributed} are unsupervised and perform well when trained on billions of text documents. 
Results from the word2vec algorithm show that in addition to capturing precise syntactic and semantic information, word embeddings obtained from these algorithms demonstrate linear structure particularly well suited for performing analogy tasks.

\textit{This paper focuses on sentiment analysis for problem domains where obtaining large amounts of data is problematic}. A typical example is that of data obtained from discussion forums that are part of digital health intervention treatments. Such treatments have demonstrated effectiveness in substance use disorders~\cite{mohr2013behavioral,moore2011computer}. Textual data obtained from these discussion forums is rich in sentiments such as determination, pleasure, anger, fear etc. The goal of this intervention treatment is to prevent relapse in users via timely intervention facilitated by human moderators and machine learning algorithms. Though forum moderators can monitor and provide support when participants are struggling, considerable labor is involved in reviewing and deciding the risk level of each text message.

By analyzing textual data for sentiment, efficient algorithms can be developed for predicting relapse. However, challenges with this data are, (i) the amount of unlabeled data is small, as the number of active users are modest, the number of posts they make in the on-line forum is modest (on the order of a few thousand) (ii) obtaining labels for this data is hard, as they need human moderated expertise to judge if a certain post is `positive/benign' which implies that the individual is doing fine or `negative/threat' implying that the individual is vulnerable and is likely to relapse soon.

The contributions of this paper are two fold. First, this paper introduces the Supervised Word Embedding for Sentiment Analysis (SWESA) algorithm (Section~\ref{optfor}). This is an iterative algorithm that 
minimizes a cost function for both a classifier and word embeddings under unit norm constraint on the word vectors. SWESA uses document labels for learning word embeddings. Using document lables overcomes the problem of small-size training data and allows learning of meaningful word embeddings. In contrast, state of the art algorithms like word2vec use large amounts of training data and learn word embeddings in an unsupervised fashion.

Second, word embeddings learned via SWESA are polarity aware as demonstrated via extensive experiments on standard data sets like Imdb, Yelp, Amazon etc (Section~\ref{exps}). For example, `Awful/Good' is the antonym pair returned via SWESA as opposed to `Awful/Could' obtained via word2vec. Such polarity aware word embeddings are suitable to perform word antonym tasks. In addition, SWESA has significant improvement over the state-of-the-art in word embeddings when used in a sentiment analysis framework.

Section 2 presents related work and Section 5 concludes this work.
\section{Related Work}
This work is related to two important areas in NLP each with a vast amount of related literature. In keeping with space constraints, this section briefly discusses major contributions from both areas.

\textbf{Word vector representations:} Earliest vector representation of words were via Vector Space Models (VSM)~\cite{turney2010frequency}. A popular example of the VSM is Latent Semantic Indexing (LSI)~\cite{deerwester1990indexing} that works on a matrix of co-occurence counts such as the term frequency-inverse document frequency (tf-idf) to learn word embeddings. Variants of the LSI involve different measures for co-occurence such as the square root of word counts~\cite{rohde2006improved}, logarithms~\cite{dumais2004latent} etc. The more recent state-of-the-art are neural network based language models that use the weights of the neural network as internal representation of a word. Neural network models are rich with initial contributions by~\cite{rumelhart1988parallel}. Successful modern incarnations of neural network models lead to the word2vec algorithm~\cite{mikolov2013efficient} which uses energy-based techniques and GloVe which uses matrix factorization techniques~\cite{pennington2014glove}. The main idea behind word2vec is to learn vector representations of words so that they maximize the probability of contiguous $c$ tuples occurring in the corpus while at the same time minimizing the probability of random $c$-tuples. Furthermore word2vec paper posits a probabilistic model based on the sum of dot products between a word and the nearby words. This model has successfully produced efficient word vector embeddings that exhibit linear properties desirable for use in applications such as word analogy tasks.
Latent variable probabilistic models~\cite{blei2003latent,blei2012probabilistic} and extensions have also been used for word embeddings.
All of the above methods learn word embeddings in an unsupervised fashion. However, using labeled data can often help with learning sentiment-aware word embeddings more appropriate to the corpus at hand. Such word embeddings can be used in sentiment analysis tasks.

\textbf{Sentiment Analysis:}
In their work~\cite{maas2011learning} propose a probabilistic model that captures semantic similarities among words across documents. This model leverages document label information to improve word vectors to better capture sentiment of the contexts in which these words occur. The probabilistic model used by is similar to that in Latent Dirichlet Allocation (LDA)~\cite{blei2003latent} in which each document is modeled as a mixture of latent topics. In~\cite{maas2011learning}, word probabilities in a document are modeled directly assuming a given topic.

A supervised neural network based model has been proposed by~\cite{tang2014learning} to classify Twitter data. The proposed algorithm learns sentiment specific word vectors, from tweets making use of emoticons in text to guide sentiment of words used in the text instead of annotated sentiment labels. 
The Recursive Neural Tensor Network (RNTN) proposed by~\cite{socher2013recursive} classifies sentiment of text of varying length. To learn sentiment from long text, this model exploits compositionality in text by converting input text into the Sentiment Treebank format with annotated sentiment labels. The Sentiment Treebank is based on a data set introduced by Pang and Lee~\cite{pang2005seeing}. This model performs particularly well on longer texts by exploiting compositionality as opposed to a regular bag of features approach.

\textbf{Notation}: Throughout this paper we shall denote word vectors as $\vw_{j} \in \bbR^{k}$, for $j=1,\hdots,V$, where $V$ indicates the size of the vocabulary. The matrix of word vectors is $\mW$ where $\mW =[\vw_{1},\vw_{2},\hdots,\vw_{V}] \in \bbR^{k \times V}$. The classifier to be learned is represented by $\vtheta \in \bbR^{k}$, weights of word vectors $\vw_{j}$ in document $i$ are contained in the vector $\vphi_{i} \in \bbR^{V}$, and the document label of the $i^{th}$ document is indicated by $y_{i}$, document $i$ is represented as $d_{i} = \mW\mb_{i}$. Let $\mg = [\mb_{1},\mb_{2},\hdots,\mb_{N}] \in \bbR^{V \times N}$ be the matrix containing weight vectors $\mb_{i}$ and vector $\vy = [y_{1},y_{2},...y_{N}]$ be the vector containing document labels.

\section{Supervised Word Vectors for Sentiment Analysis}\label{optfor}
Given a collection of documents $d_1, d_2, \ldots d_N$ with binary sentiments $y_1,y_2,\ldots, y_N$ respectively, the aim is to learn a classifier that when given a new, previously unseen document $d$ can accurately estimate the sentiment of the document. There could be class imbalance in the training data and the algorithm should explicitly account for such a class imbalance. This problem is approached by introducing a new algorithm called SWESA. SWESA simultaneously learns word vector embeddings and a classifier, by making use of document polarity/sentiment labels. Representation of documents within SWESA is motivated by the fact that in short texts like ``I am sad", ``I am happy", polarity of the sentence hinges on the words ``sad" and ``happy". As a result, by learning polarity aware word embeddings, good vector representations for documents can be achieved. For instance, in the above example,  the distance between the vectors $(\vw_{I} + \vw_{am} + \vw_{sad})$ and $(\vw_{I} + \vw_{am} + \vw_{happy})$ would capture dissimilarities in sentiment of these two documents while at the same time reflecting similarities in sentence structure.

Text documents in this framework are represented as a weighted linear combination of words in a given vocabulary. Weights can be either the term frequencies (tf) of words within each document or term frequency-inverse document frequency (tf-idf). Weights provided as input to SWESA for experiments described in Section 4 are term frequencies. This weighting scheme is chosen to mimic the concept of local context used in the word2vec family of algorithms. Global co-occurrence information can be leveraged by using tf-idf for weighting words in documents. Such an approach in not entirely unheard off in sentiment analysis tasks, where word embeddings are considered as features for a classification algorithm~\cite{labutov2013re}.

SWESA aims to find vector representations for words, and by extension of text documents such that applying a nonlinear transformation $f$ to the product $(\theta^{\top}\mW\vphi)$ results in a binary label $y$ indicating the polarity of document. Mathematically we assume that,
\begin{align}\label{meq}
\bbP[ Y = 1|d=\mW\vphi,\vtheta] = f(\vtheta^{\top}\mW\vphi)
\end{align}
for some function $f$. In order to solve for $\vtheta \text{and} \mW$, a regularized negative likelihood minimization problem is solved. This optimization problem is as~\eqref{meq} and can be solved as a minimization problem with objective function,
\begin{equation}\label{nll}
\begin{split}
J(\vtheta,\mW)\defeq \frac{-1}{N}\Bigl[\Cp\sum_{y_i = +1}\log\bbP(Y=y_{i}|\mW\mb_{i},\vtheta)\\
+\Cm_{-}\sum_{y_i = -1} \log\bbP(Y=y_{i}|\mW\mb_{i},\vtheta)\Bigr]\\
+ \lambda_{\vtheta}||\vtheta||_{2}^{2}.
\end{split}
\end{equation}
This optimization problem can now be written as
\begin{align}
\label{eq1}
\min_{\substack{\vtheta\in \bbR^k,\\\mW\in \bbR^{k\times V}}}&~~~ J(\vtheta,\mW)\\
\nonumber{}
 \text{s.t.}&~ ||\vw_{j}||_{2} = 1 ~\forall j=1,\ldots V.
 \end{align}
The vector $\vphi_i$ is a vector of weights, corresponding to the different words, for document $d_i$. As mentioned previously, for testing SWESA term frequencies of different words in a certain document $i$ are used in $\vphi_{i}$. $\lambda_{\theta}>0$ is the regularization parameter for the classifier $\theta$, $\Cp$ is the cost associated with misclassifying a document from the positive class and $\Cm_{-}$ is the cost associated with misclassifying a document from the negative class. Following the heuristic suggested by~\cite{lin2002support}, $\Cp = \frac{\nm_{-}}{N}$ and $\Cm_{-} = \frac{\np}{N}$, where $\np$ is the number of positive documents in the corpus and $\nm_{-}$ is the number of negative documents in the corpus. This scheme is particularly useful when dealing with data sets with imbalanced classes. When using a balanced data set $\Cp = \Cm_{-}$. Sentiment in a given document is captured by the document label $y_{i}$, which in this framework is a binary label that capture sentiments such as `positive/negative' or `threatening/benign' depending on the data set.

The unit norm constraint in the optimization problem shown in~\eqref{eq1} is enforced on word embeddings to discourage degenerate solutions of $\vw_{j}$. For example in the absence of this constraint, the optimal $\vw_{j}^{*}$ is typically a vector of zeros. Note that this optimization problem is bi-convex,  but it is not jointly convex in the optimization variables. Algorithm~\ref{algo} shows the algorithm that we use to solve the optimization problem in~\eqref{eq1}. This algorithm is an alternating minimization procedure that initializes the word embedding matrix $\mW$ with $\mW_0$ and then alternates between minimizing the objective function w.r.t. the weight vector $\vtheta$ and the word embeddings $\mW$. 
\begin{algorithm}
 \begin{algorithmic}[1]
 \REQUIRE $\mW_{0}$, $\mg$, $\Cp$, $\Cm_{-}$, $\lambda_{\theta}$, $0<k<V$,
 Labels: $\vy = [y_{1}, \hdots, y_{N}]$, Iterations: $T>0$,
 \STATE Initialize $\mW = \mW_{0}$.
 \FOR{$t=1,\ldots,T$}
 \STATE Solve $\vtheta_{t}\leftarrow \arg\min_{\vtheta}J(\vtheta,\mW_{t-1})$.
 \STATE Solve $\mW_{t}\leftarrow \arg\min_{\mW}J(\vtheta_t,\mW)$.
 \ENDFOR
 \STATE Return $\vtheta_{T},\mW_{T}$
 \end{algorithmic}
\caption{Supervised Word Embeddings for Sentiment Analysis (SWESA)\label{algo}}
\end{algorithm}
\subsection{Logistic regression model}
The optimization problem in~\eqref{nll} assumes a certain probability model and minimizes the negative log-likelihood under norm constraints. While, the specific goal of the user might dictate an appropriate choice of probabilistic model, for a large class of classification tasks such as sentiment analysis, the logistic regression model is widely used. In this section it is assumed that the probability model of interest is the logistic model. Under this assumption the minimization problem in Step 3 of Algorithm~\ref{algo} is a standard logistic regression problem~\footnote{A bias term, $\gamma$ can be trivially introduced in the logistic regression model.}. Many specialized solvers have been devised for this problem and in this implementation of SWESA, a standard off-the-shelf solver available in the scikit-learn package in Python is used. In order to solve the optimization problem in line 4 of Algorithm~\ref{algo} a projected stochastic gradient descent (SGD) with suffix averaging~\cite{rakhlin2011making} is used. In suffix averaging the last few iterates obtained during stochastic gradient descent are averaged. Suffix averaging guarantees that the noise in the iterates is reduced and has been shown to achieve almost optimal rates of convergence for minimization of strongly convex functions. For experiments in Section~\ref{exps} we set $\tau=50$.

Gradient updates for $\mW$ given $\vtheta$ are of the form
\begin{equation}
\begin{split}
\nabla J(\vtheta,\mW) = \frac{1}{N}\Bigl[\sum_{y_i=+1}\frac{-\Cp y_{i}\vtheta\vphi_i^\top}{1+e^{y_{i}(\vtheta^{\top}\mW\mb_{i})}}+ \\
\sum_{y_i=-1}\frac{-\Cm_{-}y_{i}\vtheta\vphi_i^\top}{1+e^{y_{i}(\vtheta^{\top}\mW\mb_{i})}}\Bigr].
\end{split}
\end{equation}
 Algorithm~\ref{SGD} implements the SGD algorithm (with stochastic gradients instead of full gradients) for solving the optimization problem in step 4 of Algorithm~\ref{algo}.
\begin{algorithm}[h]
 \begin{algorithmic}[1]
 \REQUIRE $\vtheta, \gamma, \mW_{0}$, Labels: $\vy=[y_{1},\hdots,y_{N}]$, Iterations: N,
 step size: $\eta>0$, and suffix parameter: $0<\tau\leq N$.
\STATE Randomly shuffle the dataset.
 \FOR{$t=1,\ldots,N$}
 \STATE Set $C_t = C_{+}$ if $y_t = +1$, $C_t = C_{-}$ if $y_t = -1$.
 \STATE $\widetilde{\mW}_{t+1} = \mW_{t} -  \frac{\eta C_t}{1+e^{y_{i}(\vtheta^{\top}\mW\mb_{i})}} \times (-y_{i}\vtheta\vphi_i^\top)$
 \STATE $\mW_{t+1,j} = \mW_{t+1,j}/||\mW_{t+1,j}||_2 ~\forall j =1,2,\ldots,V$
 \STATE $\eta \leftarrow \frac{\eta}{t}$
 \ENDFOR
 \STATE Return $\mW = \frac{1}{\tau}\sum_{t=N-\tau}^{N} \mW_{t}$
 \end{algorithmic}
 \caption{Stochastic Gradient Descent for $\mW$}\label{SGD}
 \end{algorithm}
\subsection{Initialization of $\mW$}
Two different initialization procedures are used to obtain $\mW_{0}$. The first method uses the Latent Semantic Analysis~\cite{dumais2004latent} procedure to form the matrix of word vectors $\mW_{0}$ from the given corpus of text documents. The second method uses the word2vec algorithm to form word vector matrix $\mW_{0}$ from the corpus.
\subsection{Dimensionality of Word Vectors}\label{dim}
In most previous literature on learning word embeddings the choice of $k$ is ad-hoc and usually fixed to some small number. In this paper, it is suggested that the spectrum of matrix $\mg$ be used to determine $k$. Typically, $k$ is required to be large enough so as to capture the intricacies in the data but at the same time small enough to avoid over fitting. In order to find the best $k$, the effective rank of the matrix $\vPhi$ is calculated. The effective rank~\cite{ganti2015matrix} of a matrix $\mg$ is defined as the smallest $k\in \bbN$, such that the best rank-k approximation, $\mg_k$, of the matrix $\mg$, satisfies
\begin{align}
\err_k(\vPhi)  \defeq \frac{||\mg - \mg_{k}||_{F}^{2}}{||\mg||_{F}^{2}} \leq \epsilon,
\end{align}
Here $||\cdot||_{F}$ indicates the Frobenius norm. This notion of effective rank has the intuitive meaning that the energy in the the $k+1, \ldots$ singular value of the matrix $\mg$ is small relative to the entire spectrum. To demonstrate that such choices of $k$ are good a simple synthetic experiment is performed. SWESA is run on a synthetic data set of 400 text documents split into 5 pairs of training and testing data sets. A polarized vocabulary of 40 words is built, comprising of 15 positive, 15 negative and 10 neutral words. A text document is assigned a negative label if at least 70$\%$ of the words in the document are negative. Similarly, a text document is labeled positive if at least  70$\%$ of the words in the document are positive. This synthetic data set is unbalanced, with 10$\%$ positive documents and the rest negative.
\begin{figure}
\centering
   \includegraphics[scale=0.23]{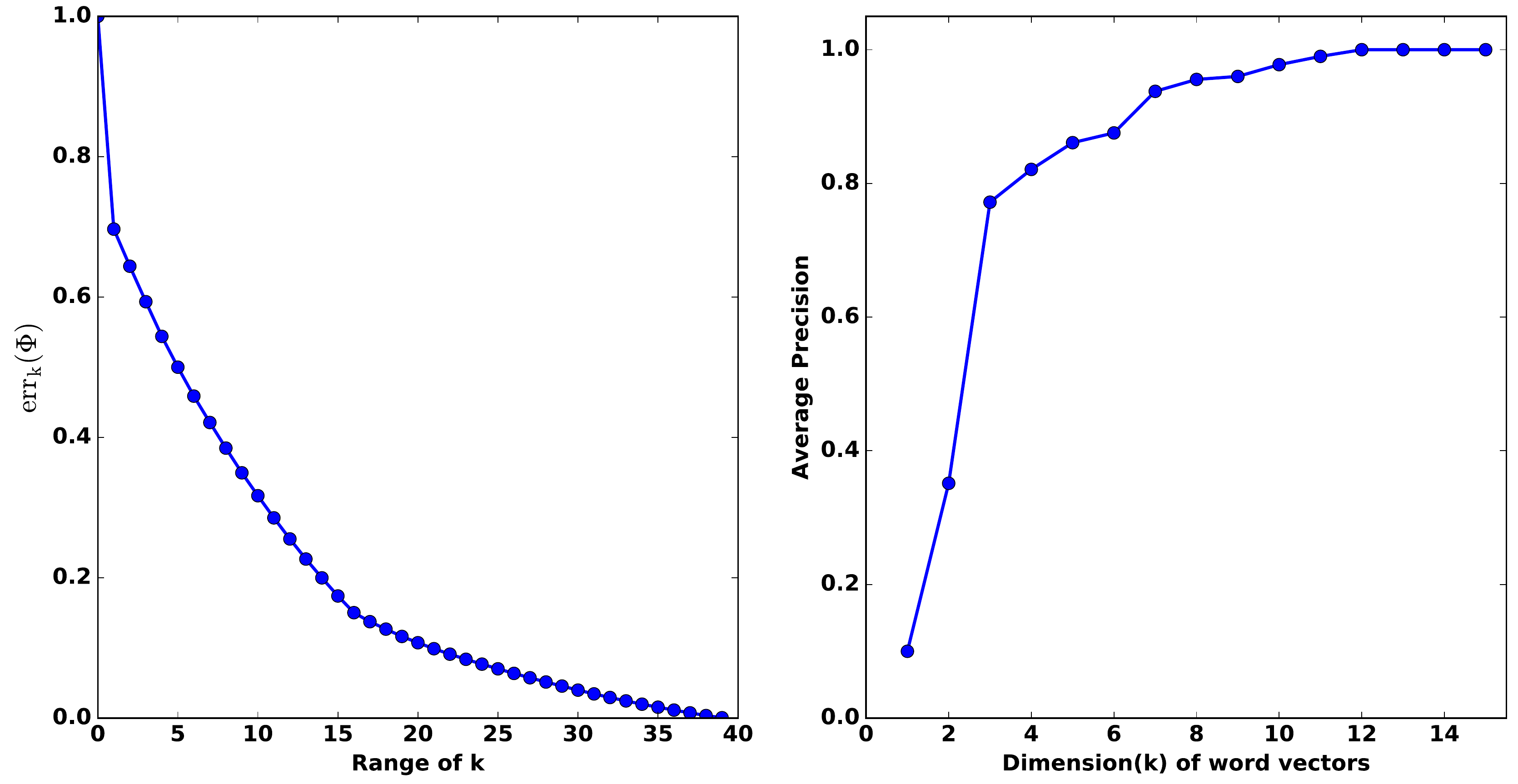}
   \caption{\textit{This figure shows $\err_k(\vPhi)$ versus $k$ on the left, and the average precision versus dimension plot for the learned word vectors on the right.}}\label{compfig}
\end{figure}
The $\mg$ matrix here is a matrix of term frequencies.
Since the data set is relatively noise free, a value of $\epsilon=0.15$. As can be seen from Figure~\ref{compfig}, the effective rank at this choice of $\epsilon$ is $k = 15$. For this value of $k$ the average precision is $1$. This demonstrates, the fact that the above definition of effective rank provides us with a good mechanism to pick good values of $k$.

\subsection{Convergence of SWESA and comparison to other algorithms}
At a high level, SWESA can be seen as a variation of the supervised dictionary learning problem (SDL). Within SDL~\cite{mairal2009supervised} given labeled data $(x_{1}, y_{1}), (x_{2}, y_{2}),\dots,(x_{n}, y_{n})$, and unlabeled part of the data that lies in a $d$ dimensional space, the goal is to learn a dictionary $D$ of size $d\times k$, $(k>>d)$ such that each $x_{i} = Dz_{i}$ where $z_{i}$ is a sparse encoding of $x_{i}$ w.r.t. dictionary $D$. Further, the label is generated by a linear classifier w.r.t $z_{i}$, i.e. $y_{i} = \mW^{\top}z_{i}$.  The learning problem is to estimate the dictionary, the codes of each data point and the classifier.
SWESA can be roughly mapped to the SDL by considering dictionary $D$ of size $k \times V$, where each column corresponds to a word embedding.

However there are three main differences between SDL and SWESA. (i) In SDL the input is a labeled dataset where each data point is already represented as a vector. This allows, a definition of reconstruction error that is used in algorithms designed for SDL. In contrast, SWESA has labeled unstructured data which does not have a direct vector representation and the aim is to learn vector representations for such data. As a result the notion of reconstruction error used in SDL does not apply to SWESA and hence the optimization formulation used is significantly different from the one used in SDL (ii) In SDL sparse encoding of each data point is to be learned, whereas in SWESA this sparse encoding is considered to be known and is proportional to the number of times a word appears in the document. (iii) Finally, in SDL the classifier is a high-dimensional vector that acts on the latent codes.
For SDL and other problems such as matrix completion~\cite{jain2013low}, convergence properties of alternating minimization have been studied. While the current analysis techniques might not apply to SWESA, due to the above mentioned differences, we conjecture that similar ideas might be useful for convergence analysis of SWESA.

Standard methods like Naive Bayes use one-hot encoding for words and hence fails to capture semantic relationships between words. In contrast, SWESA learns word embeddings that capture polarity. Neural network models learn complicated functions on the data, which makes them a poor algorithmic tool in the presence of limited data.
\section{Experimental Evaluation and Results}\label{exps}
To examine its effectiveness, SWESA is compared against the following baselines,
\begin{enumerate}
\item \textbf{Naive Bayes classifier:} The classic Naive Bayes classifier for sentiment classification based on the Bag-of-Words features, optimized in NLTK toolkit in Python is used.
\item \textbf{Recursive Neural Tensor Network (RNTN):} RNTN proposed by~\cite{socher2013recursive} learns compositionality form text of varying length and performs classification in a supervised fashion with fine grained sentiment labels. Since SWESA is aimed at binary classification, RNTN is also used in a binary classification framework. RNTN is shown to perform better than the previously proposed Recursive Auto Encoder (RAE) by~\cite{socher2011semi} and hence SWESA is not compared against RAE.
\item \textbf{Two-Step (TS):} This baseline is introduced to test the effectiveness of unsupervised embedding algorithms like LSA and word2vec as features for document sentiment classification . 
Two-step performs the following two steps to perform sentiment analysis. (i) Learn the unigram word embeddings in an unsupervised fashion and use them to obtain document embeddings via weighted linear combination. (ii) Use the obtained document embeddings to learn a logistic regression classifier for sentiment analysis.
\end{enumerate}
SWESA is compared against RNTN and not Sentiment-Specific Word Embeddings (SSEW) which is a competing neural network model developed by~\cite{tang2014learning} for three main reasons, i) the SSEW algorithm was developed specifically for sentiment analysis on twitter data and uses emoticons in the tweets as sentiment labels. In contrast in the data sets considered here emoticons are usually absent. Moreover, the structure and language characteristics of Twitter data is unlike the datasets of interest in this work, making SSEW unsuitable~\cite{blitzer2007biographies}. ii) the RNTN algorithm can handle texts of varying length. In contrast SSEW is limited to tweets which are always less than 140 characters long. iii) a well developed, readily usable code is available for RNTN, but not for SSEW.

\textbf{Experimental Set Up:} SWESA is tested against the baselines on four data sets, some of which are balanced and some of which are unbalanced.
Each data set is split into 10 train-test data pairs. In the case of the unbalanced data set the ratio of classes is held consistent across training and test data pairs. The hyperparameter $\lambda_{\theta}$ is tuned on the training data via cross validation. Similarly, $T$ i.e the number of iterations of SWESA until convergence, is determined by running the experiment on the training data for a range of values of $T$. The value of $T$ beyond which there is no significant change i.e the difference between consequent values of the objective function is $\leq 10^{-5}$, is selected. Since real data sets are noisier compared to the synthetic data set used in Section~\ref{dim}, $\epsilon = 0.3$ is selected. Average Area Under the Curve (AUC) and precision scores from all 10 test data sets are reported.
Precision (pr) is calculated as the ratio of number of true positives (tp) to the number of true positives (tp) + false positives (fp) i.e, $pr = \frac{tp}{tp+fp}$. Area Under the Curve (AUC) is obtained by applying the trapezoidal rule to calculate area from the ROC curve. All data sets used in Section~\ref{results} are tokenized and non textual characters are removed. Since the data sets are small, unlike in ~\cite{le2014distributed} all words tokenized from the data sets are retained in the vocabulary. word2vec is trained using hyperparameters similar to the default values in~\cite{le2014distributed}. Similarly, default hyperparameters reported by~\cite{socher2013recursive} are used for training the RNTN.
\subsection{Results on sentiment analysis task.}
\begin{figure*}
 \centering
 \begin{subfigure}[b]{0.49\textwidth}
 \includegraphics[width=\textwidth]{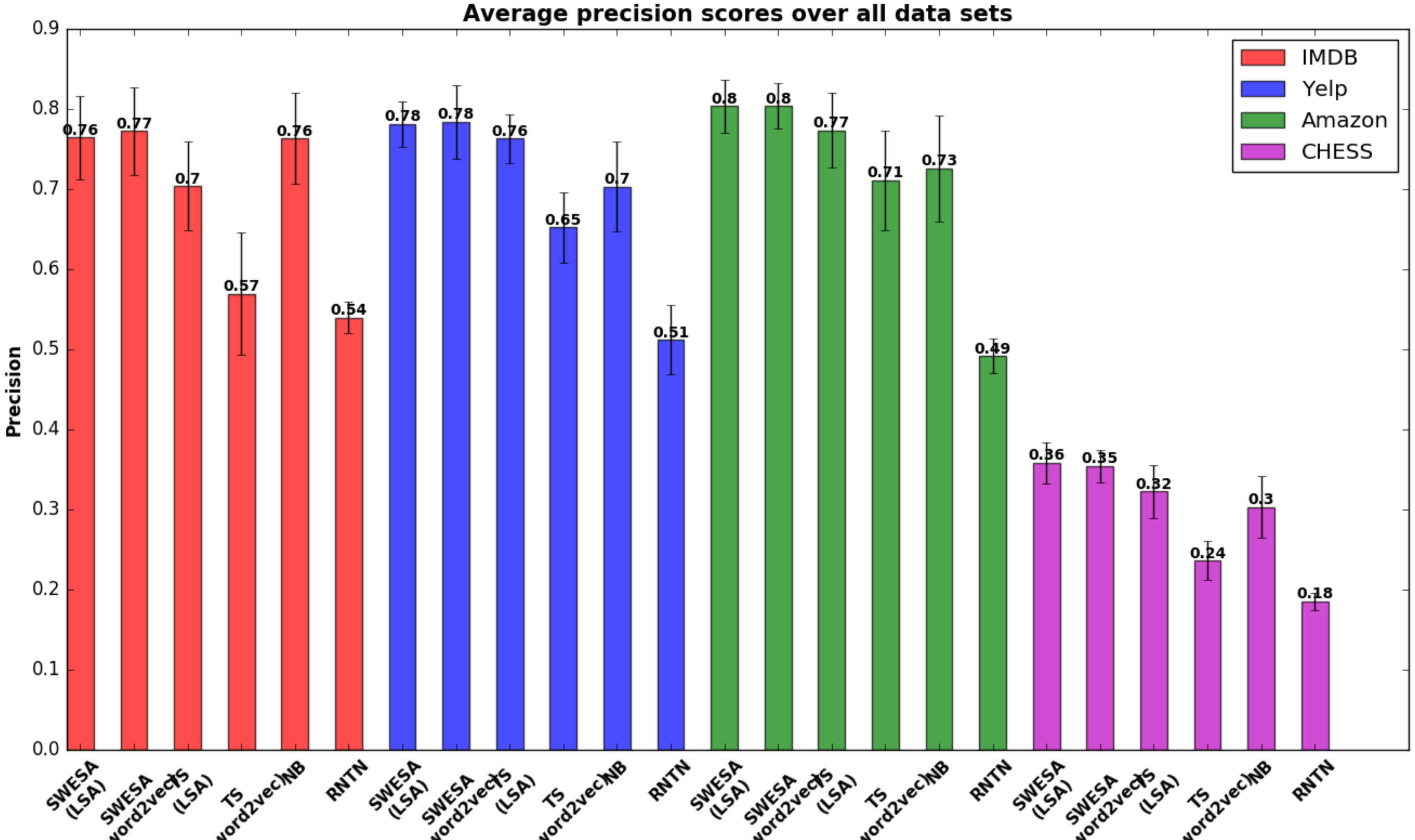}
 \caption{\textit{This figure shows the average precision scores obtained by baselines and SWESA on all four data sets. Each error bar represents the average precision score obtained by running all algorithms on 10 testing sets.}}\label{pres}
 \end{subfigure}
 \begin{subfigure}[b]{0.485\textwidth}
 \includegraphics[width=\textwidth]{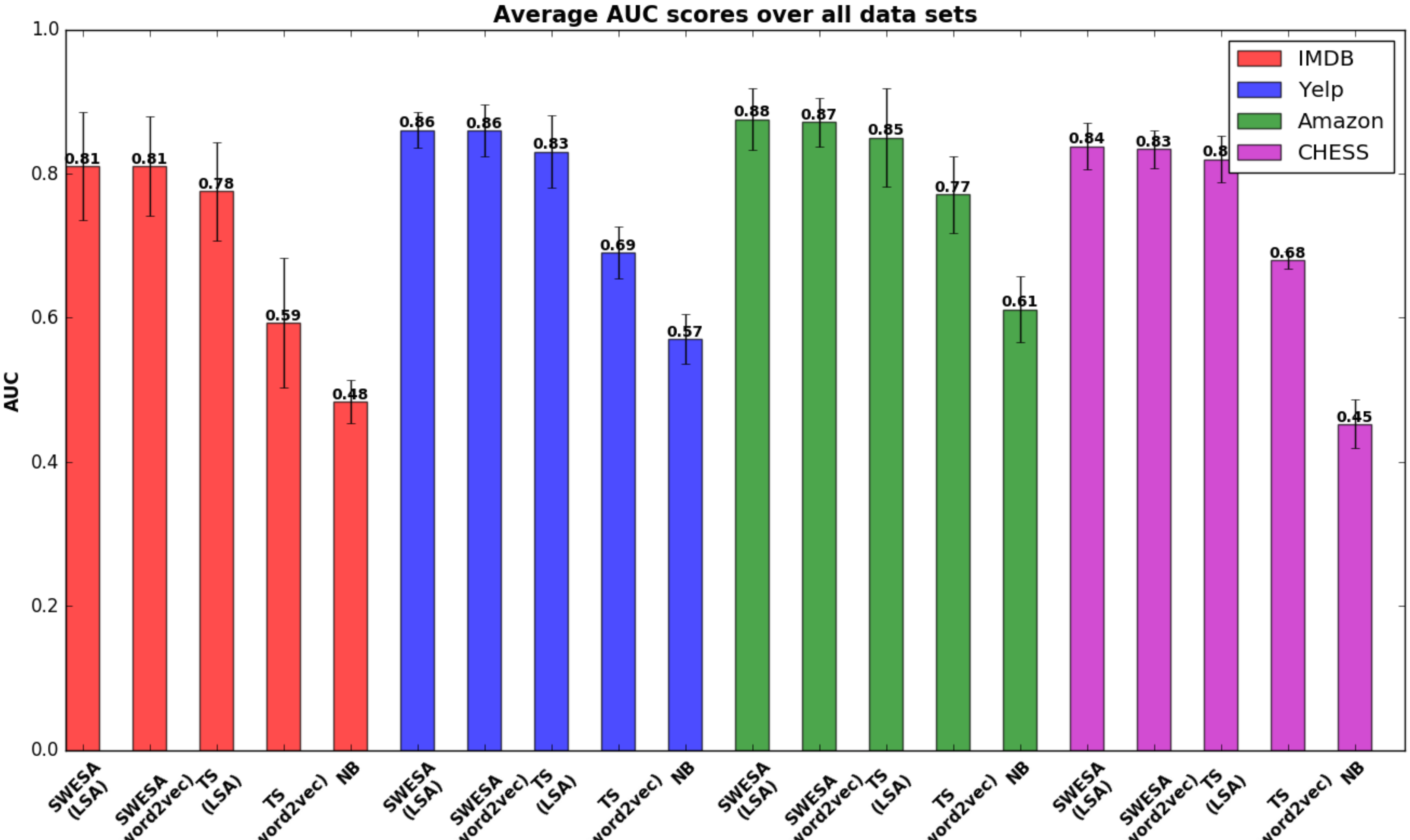}
 \caption{\textit{This figure shows the average AUC scores obtained by baselines and SWESA on all four data sets. Each error bar represents the average AUC score obtained by running all algorithms on 10 testing sets.}}\label{auc}
 \end{subfigure}
\end{figure*}

{\label{results}}
Figures~\ref{pres} and~\ref{auc} show the average precision and average AUC\footnote{AUC scores are not available for RNTN since it is not possible to determine prediction probabilities from this model.} scores respectively of all baselines and SWESA on four data sets of which three balanced data sets (Yelp, Amazon and IMDB) consist of 1000 reviews of food, products and movies respectively. Each review is labeled as `Positive' or `Negative'. These data sets are available for download from the UCI repository~\cite{Lichman:2013}. The CHESS data set consists of 2500 documents obtained from a mobile phone based intervention treatment that provides services for recovery maintenance and relapse prediction for alcohol addicts~\cite{gustafson2014smartphone}. This is an unbalanced data set where the number of documents suggestive of relapse ($\approx 8\%$) in a user are far outnumbered by users discussing their sobriety. Each message is labeled as `threat' suggesting a relapse risk and `not threat' indicating well being. This data set is proprietary of the study conducted by~\cite{gustafson2014smartphone}.

\textit{Both the neural network based baselines RNTN (sentiment classification) and word2vec (word embeddings) based Two-Step baseline perform weakly as opposed to SWESA and other baselines.} This observation is consistent with behavior of neural network based algorithms on small data sets. Despite being pre trained on the Wikipedia corpus, word2vec derived embeddings used in Two-Step fail to perform as well as SWESA consistently on all four data sets, achieving a maximum precision of 0.7109 on the Amazon data set. On the same data set the two different initializations of SWESA achieve precision scores of 0.8036 and 0.8031. Failure of Two-step with word2vec can be attributed to the lack of supervision during training and also to the disparity in training and test data.

SWESA learns meaningful embeddings from text as opposed to methods like Naive Bayes where word frequencies are used to obtain one hot encodings for documents. Hence embeddings learned via SWESA are better suited for sentiment analysis. This can be seen via the average precision and AUC of 0.7254 and 0.6116 achieved by NB on the Amazon data set as opposed to the average precision and AUC of 0.8031 and 0.8754 achieved by SWESA with the LSA initialization. As seen from figures~\ref{pres},\ref{auc} this behavior is consistent across the other balanced data sets and the CHESS data set with imbalanced classes.
To highlight the qualitative performance of SWESA, cosine similarity between document representations via SWESA and Two-Step are evaluated. Top three reviews obtained from Two-Step and SWESA most similar to the sample review \textit{``First off the reception sucks, I have never had more than 2 bars, ever."} are,
\begin{enumerate}
\item SWESA \begin{itemize}
            \item ``The worst phone I've ever had.... Only had it for a few months."
            \item ``I recently had problems where I could not stay connected for more than 10 minutes before being disconnected."
            \item ``Then I exchanged for the same phone, even that had the same problem."
            \end{itemize}
\item Two-Step \begin{itemize}
               \item ``But it does get better reception and clarity than any phone I've had before."
               \item ``none of the new ones have ever quite worked properly "
               \item ``In the span of an hour, I had two people exclaim ``Whoa - is that the new phone on TV?!?""
               \end{itemize}
\end{enumerate}
A similar analysis is performed and holds consistently across the other 3 data sets and is available in the supplemental material. This shows that SWESA propagates document level polarity onto word embeddings which helps in sentiment analysis.

\textbf{Failure of pre-trained RNTN}: Neural network based RNTNs work well when trained on large data sets. In their work~\cite{socher2013recursive} train RNTN on the Pang and Lee data set~\cite{pang2005seeing} of 10k movie reviews. Table shows the average precision obtained by pre-trained RNTN on all data sets. Note that difference in average precision scores between pre-trained RNTN and SWESA is considerably small given that pre-trained RNTN is trained on a dataset that is $10$ times the size of training data for SWESA. This observation is best illustrated on the Amazon dataset where average precision of pre-trained RNTN are approximately $0.83$ and of SWESA is $0.81$.
\begin{table}\label{tab}
\begin{tabular}{|c|c|c|}
\hline
&Average Precision&STD\\
\hline
Amazon&0.8284&0.0067\\
\hline
IMDB&0.8388&0.0070\\
\hline
Yelp&0.8331&0.0111\\
\hline
\end{tabular}
\caption{\textit{This table shows the average precision obtained by pre-trained RNTN on three balanced data sets.}}
\end{table}
However, pre-trained RNTN does particularly poorly on the CHESS data set. While it is know that difference in language structure and vocabulary of training and test data introduces some error~\cite{blitzer2007biographies}, pre-trained RNTN fails to classify most messages in the CHESS data set. Also, pre-trained RNTN does a poor job in accounting for class imbalance in the CHESS data set because of which precision scores in the messages that do get classified are extremely low.

\begin{figure}[t]
\centering
 \includegraphics[scale=0.245]{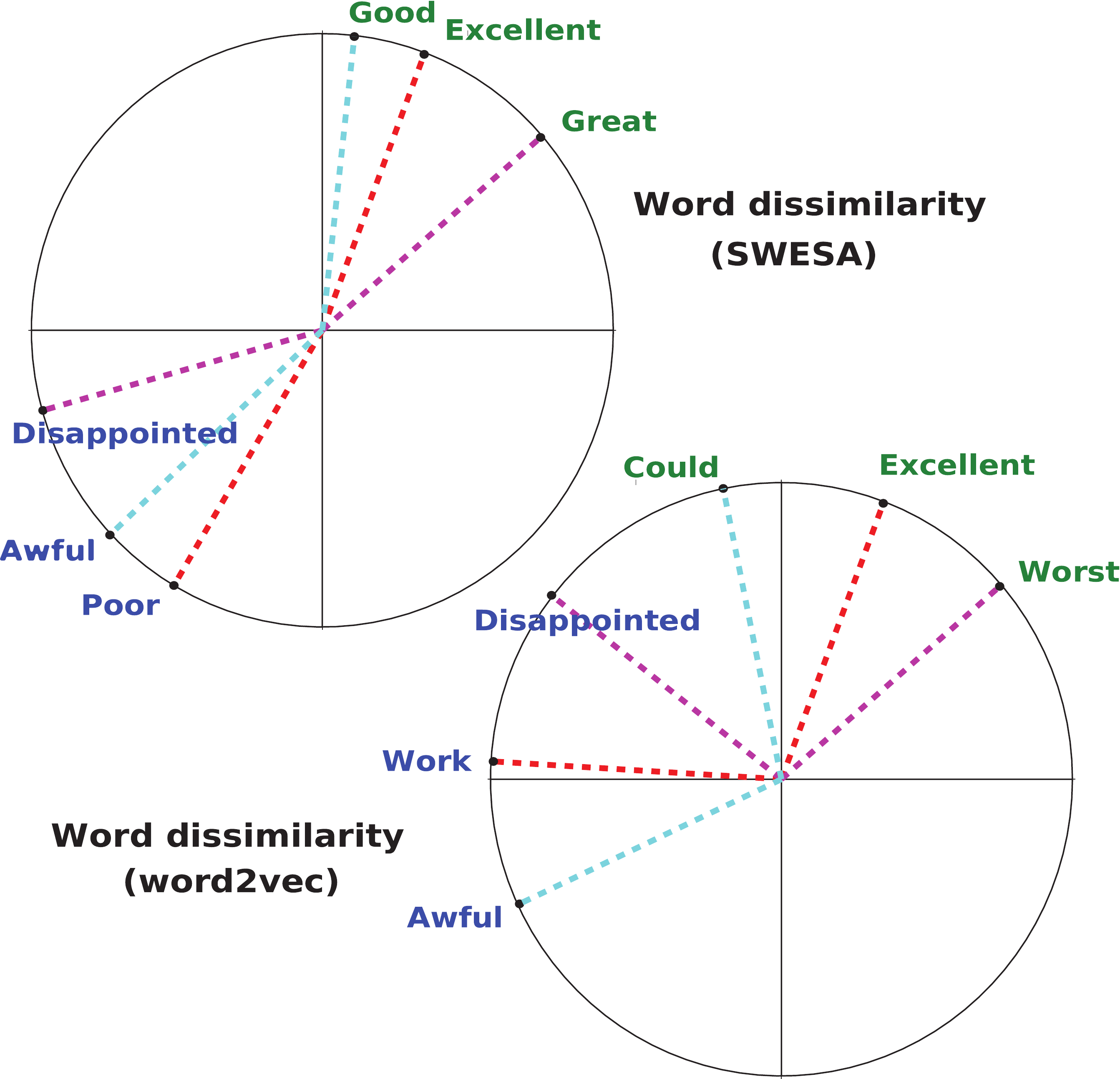}
 \caption{\textit{This figure depicts word embeddings on a unit circle. Most dissimilar word pairs are plotted based on the cosine angle between the respective word embeddings learned via SWESA and word2vec.}}\label{f2}
\end{figure}

\textbf{Polarity of word embeddings.}
The objective of SWESA is to perform effective sentiment analysis by learning embeddings from text documents with sentiment labels. As a consequence of which word level polarity is preserved in vector space. That is, given words `Good,'`fair' and `Awful,' the antonym pair `Good/Awful' is determined by calculating the cosine similarity between $\vw_{Good}$ and $\vw_{Awful}$.
Figure~\ref{f2} shows a small sample of word embeddings learned on the Amazon data set by SWESA and word2vec. The cosine similarity (angle) between the most dissimilar words is calculated and owing to the assumptions on word embeddings, words are depicted as points on the unit circle. From figure~\ref{f2} it is evident that a supervised algorithm like SWESA projects document level polarity onto word level embeddings while an unsupervised algorithm like word2vec that learns embeddings of words via virtue of word co-occurrences will fail to embed polarity.
It is important to notice that SWESA learns word polarities by using document polarities, and these word polarities are useful for antonym tasks. Unlike classical antonym tasks where examples of known antonym pairs are provided, in our setup no such pairs are provided, and yet SWESA was able to do a good job discovering antonym pairs.
For example the most dissimilar word to given word `Excellent' is `Poor' when learned via SWESA as opposed to `Work' when learned via word2vec.
Thus, word antonym pairs $(w_{a},w)$ can be obtained by calculating cosine similarities. These examples illustrate that SWESA captures sentiment polarity at word embedding level despite limited data.
\section{Conclusions and Future work}
This paper introduces SWESA, a novel iterative algorithm that simultaneously learns polarity aware word embeddings and a classifier to perform sentiment analysis in a supervised learning framework. SWESA overcomes the limitations posed by small sized data sets to neural network based learning algorithms.
Assumptions on the structure of word embeddings within SWESA preserve structural properties desirable of embeddings typically obtained via neural network based embedding algorithms. As future work, it is proposed that the geometric interpretation of the semantic relationships between word embeddings be used to mine additional semantic relationships between words and concepts in data.
\bibliography{sentbib}
\bibliographystyle{ijcnlp2017}

\end{document}